\documentclass[preprint,12pt]{elsarticle}

\usepackage{amssymb}
\usepackage{multirow}
\usepackage{algorithm}
\usepackage{algpseudocode}
\usepackage{graphicx}
\usepackage{textcomp}
\usepackage{xcolor}
\usepackage{graphics}
\usepackage{MnSymbol}
\usepackage[font=scriptsize]{caption}
\usepackage{rotating}
\usepackage{array}

\newcolumntype{P}[1]{>{\centering\arraybackslash}p{#1}}
\newcolumntype{M}[1]{>{\centering\arraybackslash}m{#1}}

\newcommand{\changetable}[1]{\color{black}{#1}}

\newcommand{\always}{\square}
\newcommand{\eventually}{\lozenge}
\newcommand{\until}{\mathcal{U}_I}

\newcommand{\ew}{\boxbox_{\ra}}
\newcommand{\sw}{\diamonddiamond_{\ra}}
\newcommand{\ag}{\mathcal{A}_{\ra}}
\newcommand{\ct}{\mathcal{C}_{\ra}}

\newcommand{\op}{\mathrm{op}}

\newcommand{\avg}{\mathrm{avg}}

\newcommand{\nb}{L_{\ra}}
\newcommand{\nbx}{\alpha_{\ra}^x(\omega, t, l)}

\newcommand{\ra}{\mathcal{D}}

\definecolor{brilliantlavender}{rgb}{0.96, 0.73, 1.0}
\definecolor{blond}{rgb}{0.98, 0.94, 0.75}
\definecolor{celadon}{rgb}{0.67, 0.88, 0.69}
\definecolor{columbiablue}{rgb}{0.61, 0.87, 1.0}
\definecolor{lavenderblush}{rgb}{1.0, 0.94, 0.96}
\definecolor{electriclavender}{rgb}{0.96, 0.73, 1.0}

\journal{Pervasive and Mobile Computing}

\newcommand{\newcontent}[1]{\textcolor{black}{#1}}
\newcommand{\revision}[1]{\textcolor{blue}{#1}}
\begin{document}

\begin{frontmatter}




\title{CitySpec with Shield: A Secure Intelligent Assistant for Requirement Formalization}


\author[inst1]{Zirong Chen}

\affiliation[inst1]{organization={Vanderbilt University},
            city={Nashville},
            state={Tennessee},
            country={USA}}

\author[inst2]{Isaac Li}

\affiliation[inst2]{organization={University of Virginia},
            city={Charlottesville},
            state={Virginia},
            country={USA}}

\author[inst3]{Haoxiang Zhang}

\affiliation[inst3]{organization={Columbia University},
            city={New York},
            state={New York},
            country={USA}}

\author[inst4]{Sarah Preum}

\affiliation[inst4]{organization={Dartmouth College},
            city={Hanover},
            state={New Hampshire},
            country={USA}}

\author[inst2]{John A. Stankovic}

\author[inst1]{Meiyi Ma}

\begin{abstract}

\newcontent{An increasing number of monitoring systems have been developed in smart cities to ensure that a city’s real-time operations satisfy safety and performance requirements. However, many existing city requirements are written in English with missing, inaccurate, or ambiguous information. There is a high demand for assisting city policymakers in converting human-specified requirements to machine-understandable formal specifications for monitoring systems. To tackle this limitation, we build \textit{CitySpec} \cite{9821044}, the first intelligent assistant system for requirement specification in smart cities. To create CitySpec, we first collect over 1,500 real-world city requirements across different domains (e.g., transportation and energy) from over 100 cities and extract city-specific knowledge to generate a dataset of city vocabulary with 3,061 words. We also build a translation model and enhance it through requirement synthesis and develop a novel online learning framework with shielded validation. The evaluation results on real-world city requirements show that CitySpec increases the sentence-level accuracy of requirement specification from 59.02\% to 86.64\%, and has strong adaptability to a new city and a new domain (e.g., the F1 score for requirements in Seattle increases from 77.6\% to 93.75\% with online learning). After the enhancement from the shield function, CitySpec is now immune to most known textual adversarial inputs (e.g., the attack success rate of DeepWordBug \cite{gao2018black} after the shield function is reduced to 0\% from 82.73\%). We test the CitySpec with 18 participants from different domains. CitySpec shows its strong usability and adaptability to different domains, and also its robustness to malicious inputs.}

\end{abstract}




\end{frontmatter}


\section{Introduction}
\label{sec:introduction}

\begin{figure}[t]
    \centering
    \includegraphics[width=0.98\textwidth]{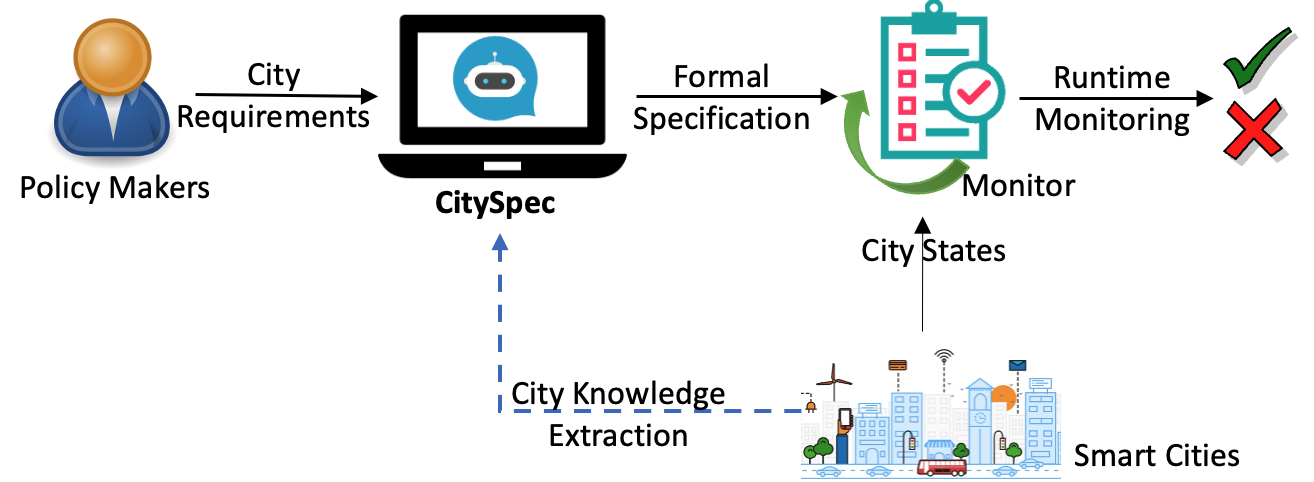}
    \caption{CitySpec in Smart Cities}
    \label{fig:CitySpec_City}
\end{figure}

With the increasing demand for safety guarantees in smart cities, significant research efforts have been spent toward how to ensure that a city’s real-time operations satisfy safety and performance requirements~\cite{ma2021toward}. Monitoring systems, such as SaSTL runtime monitoring~\cite{ma2021novel}, CityResolver~\cite{ma2018cityresolver}, and STL-U predictive monitoring~\cite{ma2021predictive}, have been developed in smart cities. Figure \ref{fig:CitySpec_City} shows a general framework of monitoring systems in smart cities. These systems are designed to execute in city centers and to support decision-making based on the verification results of real-time sensing data about city-states (such as traffic and air pollution). If the monitor detects a requirement violation, the city operators can take actions to change the states, such as improving air quality, sending alarms to police, calling an ambulance, etc.

The monitor systems have two important inputs, i.e., the real-time data streams and formal specified requirements. Despite that extensive research efforts have been spent toward improving the expressiveness of specification languages and efficiency of the monitoring algorithms, the research challenge of how to convert human-specified requirements to machine-understandable formal specifications has received only scant attention. 
Moreover, our study (see Section \ref{sec:motivation}) on over 1,500 real-world city requirements across different domains\footnote{In this paper, we define a domain as an application area in smart cities, such as transportation, energy, and public safety.} shows that,  first, existing city requirements are often defined with missing information or ambiguous description, e.g., no location information, using words like nearby, or close to. They are not precise enough to be converted to a formal specification or monitored in a city directly without clarifications by policy makers. Secondly,
the language difference between English specified requirements and formalized specifications is significant. Without expertise in formal languages, it is extremely difficult or impossible for policy makers to write or convert their requirements to formal specifications. Therefore, there is an urgent demand for an intelligent system to support policy makers for requirement specifications in smart cities.   

Despite the prevalence of developing models to translate the natural language to machine languages in various applications, such as Bash commands~\cite{fu2021transformer}, Seq2SQL~\cite{zhong2017seq2sql}, and Python~\cite{chen2021evaluating}, it is very \textbf{challenging} to develop such an intelligent system for requirement specification in smart cities for the following reasons. First, unlike the above translation tasks with thousands or even millions of samples in a dataset, there is barely any requirement specification data. 
As a result, traditional language models are not sufficient to be applied directly. Moreover, the requirements usually contain city domain-specific descriptions and patterns that existing pre-trained embeddings like BERT or GloVE cannot handle effectively. 
Furthermore, requirements from different domains and cities vary significantly and evolve over time, thus building a system that can adapt to new domains at runtime is an open research question. Good adaptability can increase user experience (e.g., policy makers do not have to clarify new terms repeatedly), while one of the major challenges is validating and filtering the new knowledge and avoiding adversarial examples online.


\newcontent{Adversarial samples can be introduced among that new knowledge if there is any malicious behavior involved in off-guard online learning. Those adversarial samples will poison the dataset used for continuous learning and system adaptation if there is no security enhancement. Here we emphasize the importance of an effective and safe validating function against malicious inputs. After reviewing the most recent literature, we find adversarial examples perturb textual inputs not only at the character level but also at a word or even sequence level. By only looking at textual information, the validation mechanism alone is not enough to protect the dataset from being poisoned or even vulnerable to malicious attacks. Because in this continuous learning scenario, the attacker could keep attacking until the model prediction changes. Thus, an effective and comprehensive approach must first detect malicious behaviors. By doing so, the validation model is kept safe and can further help protect and enrich the dataset.}      

In this paper, we target the above technical challenges and {develop \textit{CitySpec}}, an intelligent assistant system for requirement specification in smart cities. To the best of our knowledge, it is the first specification system helping city policy makers specify and translate their requirements into formal specifications automatically. As shown in Figure \ref{fig:CitySpec_City}, CitySpec is designed to bridge the gap between city policy makers and monitoring systems. It enables policy makers to define their requirements by detecting missing, inaccurate, or ambiguous information through an intelligent assistant interface. To effectively train the translation model using a small amount of city requirement data, CitySpec extracts city knowledge and enhances the learning process through requirement synthesis. CitySpec can easily adapt to a new city or application domain through online learning and validation \newcontent{under the protection of a shield model}.

\textbf{Contributions}. We summarize the major contributions of this paper as follows: 
\begin{itemize}
    \item We collect and annotate over 1,500 real-world city requirements from over 100 cities across different domains. We extract city-specific knowledge and build a dataset of city vocabulary with 3,061 words in 5 categories. 
    \item We create an intelligent assistant system for requirement specification in smart cities. 
    In the system, we build a translation model and enhance it through requirement synthesis, and 
     develop a novel online learning framework with validation under uncertainty. 
    \item We evaluate CitySpec extensively using real-world city requirements. The evaluation results show that CitySpec is effective on supporting policy makers accurately writing and refining their requirements. It increases the sentence level accuracy of requirement specification from 59.02\% to 86.64\% through city knowledge injection. It shows strong adaptability (user experience) to a new city (e.g., F1 score in Seattle from 77.6\% to 93.75\%) and a new domain (e.g., F1 score in security domain from 62.93\% to 93.95\%). 
    \newcontent{\item We overview 12 different adversarial text attacks from the most recent literature and launch them to the validation function. Based on generated adversarial samples, we develop a shield model that is hard to attack and immune to most adversarial samples. The evaluation results show its ability to significantly reduce attack success rate (e.g., the attack success rate of BertAttack is reduced from 94.00\% to 4.31\%).}
    \newcontent{\item We conducted a real user case study with 18 participants with different backgrounds. The study shows the high usability and adaptability of CitySpec not only in smart city scenarios but also in unseen domains (e.g., high user experience scores and low numbers of interactions are reported in domains like Medical and Environmental Engineering). Furthermore, the effectiveness of the shield function is shown during the survey (e.g., 87.23\% defense success rate against commonly seen adversarial attacks).}
\end{itemize}




\newcontent{This paper is an extension of \cite{9821044}. We extend with the following new contributions. First, we carefully study potential adversarial text generation. We summarize the characteristics of each attack and its effects on our system. Second, we also implement and further experiment with those 12 attacks. We find potential vulnerabilities in the initial validation model under those attacks if off-guard. Third, by targeting adversarial attacks, we enhance our system security by developing an effective and secure shield model to protect the validation model from malicious attacks. Fourth, we further conduct a comprehensive evaluation of the newly developed shield layers of CitySpec and show that CitySpec can significantly reduce the success rate of 12 types of attacks. Last, we conduct a user case study with 18 participants with different backgrounds to test the usability and adaptability of CitySpec in the smart city. Leveraging the different backgrounds of participants, we also prove the capability of CitySpec to learn continuously and adapt efficiently in unseen domains.}

\textbf{Paper organization}: In the rest of the paper, we describe the motivating study of city requirement specification  in Section \ref{sec:motivation}, provide an overview of CitySpec in Section \ref{sec:overview}, and present the technical details in Section \ref{sec:method}. We then present the evaluation results in Section \ref{sec:eval}, discuss the related work in Section \ref{sec:related} and draw conclusions in Section \ref{sec:summary}.


%
\section{Motivating Study}
\label{sec:motivation}

\begin{table*}[t]
\caption{Comparison between English requirements and formal specifications (DLD: Damerau–Levenshtein Distance)}
\scriptsize
\centering{%
\begin{tabular}{|c|p{11.3cm}|c|}
\hline
ID                 & Requirement vs Specification                                                                                                                 & DLD                 \\ \hline
\multirow{2}{*}{1} & Req: Sliding glass doors shall have an  air infiltration rate of no more than 0.3 cfm per square foot.                                            & \multirow{2}{*}{59} \\ \cline{2-2}
                   & Spec: $\mathsf{Always}_{[0, +\infty)} ~ ( \mathsf{{Sliding ~ glass ~ doors}} ~ \mathsf{air~infiltration~rate} ~ \leq 0.3 ~ \mathsf{cfm/foot}^{2})$ &                     \\ \hline
\multirow{2}{*}{2} & Req: The operation of a Golf Cart upon a Golf Cart Path shall be restricted to a maximum speed of 15 miles per hour.                              & \multirow{2}{*}{67} \\ \cline{2-2}
                   & Spec: $\mathsf{Everywhere}(\mathsf{Golf~ Cart ~Path})(\mathsf{Always}_{[0, +\infty)}(\mathsf{Golf ~Cart ~speed} < 15 ~\mathsf{miles/hour}))$       &                     \\ \hline
\multirow{2}{*}{3} & Req: Up to four vending vehicles may dispense merchandise in any given city block at any time.                                                    & \multirow{2}{*}{75} \\ \cline{2-2}
                   & Spec: $\mathsf{Everywhere}(\mathsf{city ~block})(\mathsf{Always}_{[0, +\infty)}(\mathsf{vending ~vehicles}  \leq 4))$                              &                     \\ \hline
\end{tabular}
}
\label{tab:eng_formal}
\vspace{-0.3cm}
\end{table*}


In this section, we study real-world city requirements
and their formal specification as motivating examples to discuss the demand and challenges of developing an intelligent assistant system for requirement specification in smart cities. 
\revision{We collect and annotate over 1,500 real-world city requirements (e.g., standards, codes of ordinances, laws, regulations, etc. \cite{r1, r2, r3, r4, r5, r6, r7, r8, r9, r10, r11, dcair, nashcode, memphiscode, atlcode}) from over 100 cities and regions (e.g. New York
City, San Francisco, Chicago, Washington D.C., Beijing, etc.) around the world. Those requirements also cover different application domains like transportation, environment, security, public safety, indoor environments, etc.
We make the following observations from the analysis of the requirement dataset. }

\textit{Existing city requirements are often defined with missing information or ambiguous description.} In \cite{ma2021novel}, the authors define essential elements for monitoring a city requirement.   
Within the 1500 requirements, many requirements have one or more missing elements. For example, 27.6\% of the requirements do not have location information, 29.1\% of the requirements do not have a proper quantifier, and 90\% of the requirements do not have or only have a default time (e.g., always) defined. Additionally, requirements often have ambiguous descriptions that are difficult to be noticed by policy makers. For example, a location is specified as ``nearby'' or ``close to''. As a result, it is very difficult or impossible for the monitoring system to monitor these requirements properly. It indicates a high demand for an intelligent assistant system to support the policy makers to refine the requirements.


\textit{The language difference between English specified requirements and their formal specifications is significant.}
In Table \ref{tab:eng_formal}, we give three examples of city requirements in English, their formal specification in SaSTL, and the Damerau–Levenshtein Distance (DLD)~\cite{damerau1964technique} between each pair of requirements. DLD measures the edit distance between two sequences. 
It shows that 
natural languages are different from machine-compatible input languages. Formal specifications usually consist of mathematical symbols, which makes the conversion even more difficult. 
As shown in Table \ref{tab:eng_formal}, the average DLD from English requirements to formal specifications is 67, which means that it requires an average of 67 edits. 
As a reference, the average DLD brought by translating these three English requirements to Latin is 64.67. It indicates that the conversion from English requirements to formal specifications even requires more edits than the translation of these requirements from English to Latin. In general, building a translator from English to Latin would require millions of samples. However, as an under-exploited area, there is a very limited number of well-defined requirements. Moreover, annotation of formal specifications requires specialties in formal methods and is extremely time-consuming. It presents major challenges for building such a translation model.

\section{System Overview}
\label{sec:overview}

\begin{figure*}[t]
    \centering
    \includegraphics[width=0.95\textwidth]{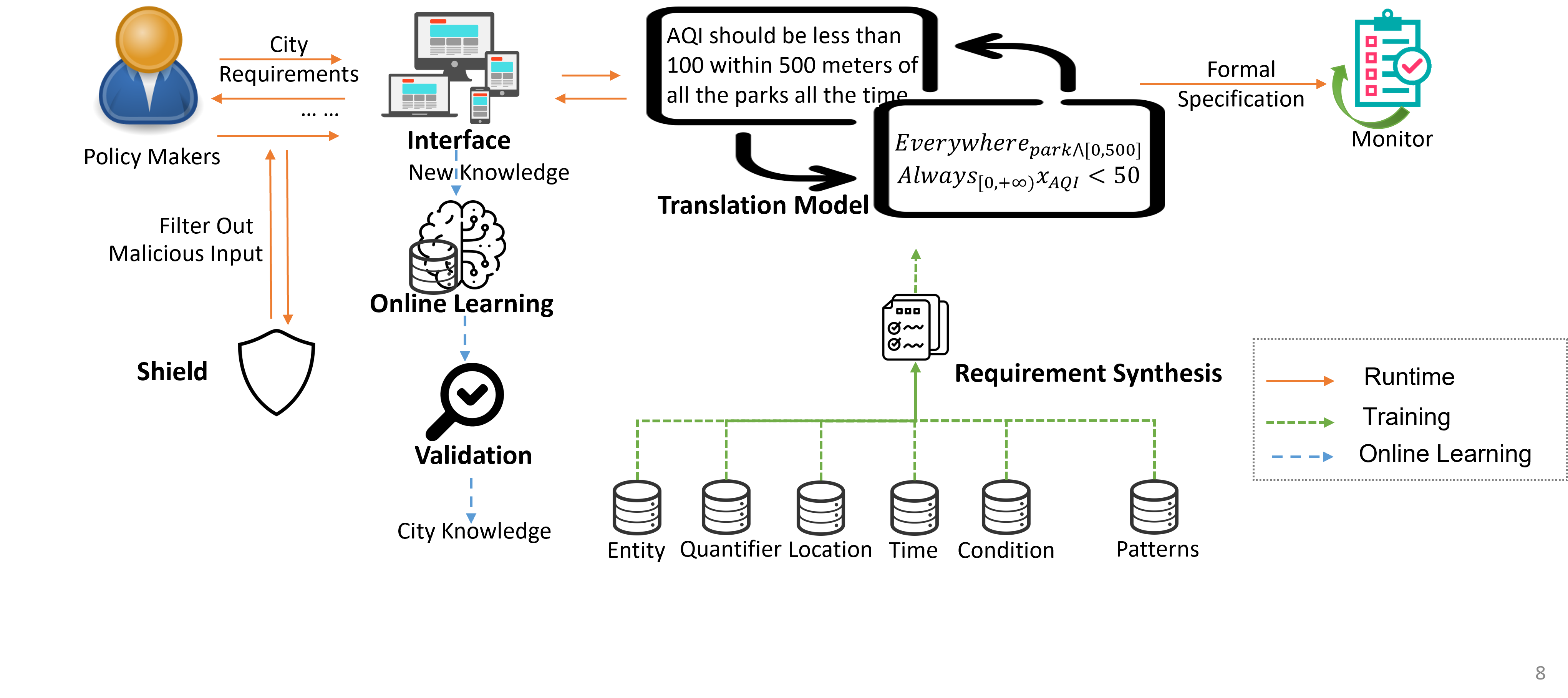}
    \caption{System Overview}
    \label{fig:Overview}
    \vspace{-0.5cm}
\end{figure*}

CitySpec is designed to bridge the gap between city policy makers and monitoring systems. It supports policy makers to precisely write city requirements in English through an intelligent interface, and then converts them to formal specifications automatically. An overview of CitySpec is shown in Figure \ref{fig:Overview}. There are four major components in CitySpec, including an intelligent assistant \textit{Interface} to communicate with policy makers (see Section \ref{subsec:interface}), a \textit{Requirement Synthesis} component to extract city knowledge and synthesize new requirements to build the translation model (see Section \ref{subsec:requirement}), a \textit{Translation Model} to convert city requirements to formal specifications (see Section \ref{subsec:translation}),  and an \textit{Online Learning} component to adapt the system to new knowledge (see Section \ref{subsec:online}). 

At runtime (as indicated by the orange arrows in Figure \ref{fig:Overview}), 
a city policy maker inputs a requirement in English through an intelligent assistant {interface}, which sends the requirements to the translation model. The \textit{translation model} converts the requirements to a formal specification and checks if there is any missing information or ambiguous description. 
The translation model is built with injected city knowledge through requirement synthesis at the training time and enhanced through online learning at runtime. 
Next, based on the returned results from the translation model, the intelligent interface communicates with the policy maker to acquire or clarify the essential information. In this process, the assistant supports the policy maker to refine the requirement until it is precisely defined and accepted by the monitor. We present the technical details in Section \ref{sec:method}, 
and develop a prototype tool of the CitySpec system and deploy it online.

\section{Methodology}
\label{sec:method}
In this section, we present the major components in CitySpec (as shown in Figure \ref{fig:Overview}). We first introduce requirement specification using Spatial-aggregation Signal Temporal Logic (SaSTL)~\cite{ma2021novel}.  Then we show the design and technical details of the intelligent assistant interface, requirement synthesis, translation model, and online learning, respectively.

\subsection{Requirement Specification using SaSTL}

SaSTL is a powerful formal specification language for Cyber-Physical Systems. We select it as our specification language because of its advantages of expressiveness and monitoring for smart cities. However, CitySpec is general and can work with other specification languages.
SaSTL is defined on a multi-dimensional \emph{spatial-temporal signal} as $\omega: \mathbb{T} \times L \to \{{\mathbb{R}\cup\{\bot\}\}} ^n$,
where $\mathbb{T}=\mathbb{R}_{\ge 0}$, represents the continuous time and $L$ is the set of locations. $X= \{x_1, \cdots, x_n \}$ is denoted by the set of 
variables for each location. 
The spatial domain $\ra$ is defined as, $\ra :=  ([d_1, d_2],\psi)$, $\psi := \top\;|\;p\;|\;\neg\;\psi\;|\;\psi\;\vee\;\psi$,
where $[d_1, d_2]$ defines a spatial interval with $d_1 < d_2$ and $d_1,d_2 \in \mathbb{R}$, and $\psi$ specifies the property 
over the set of propositions
that must hold in each location.

The syntax of SaSTL is given by 
\begin{equation*} 
\begin{array}{cl}
\varphi  :=& x \sim c\;| \
 \neg \varphi \;| \
\varphi_1 \land \varphi_2 \;|\
\varphi_1 \until \varphi_2 \;|\
\ag^{\op} x \sim c \;|\
\ct^{\op} \varphi \sim c \\
\end{array}
\end{equation*}

where $x \in X$, $\sim \in \{<, \le\}$, $c \in \mathbb{R}$ is a constant, 
$I \subseteq \mathbb{R}_{> 0}$ is a real positive dense time interval, 
$\until$ is the \emph{bounded until} temporal operators from STL. 
The \emph{always} (denoted $\always$) and \emph{eventually} (denoted $\eventually$) temporal operators can be derived the same way as in STL, where $\eventually \varphi \equiv \mathsf{true} \ \until \varphi$, and $\always \varphi \equiv \neg \eventually \neg \varphi$.
Spatial \emph{aggregation} operators $\ag^{\op} x \sim c$ for 
$\op \in \{\max, \min, \mathrm{sum}, \avg\}$ evaluates the aggregated product of traces $\op(\nbx)$ over a set of locations $l \in \nb^l$, and \emph{counting} operators $\ct^{\op} \varphi \sim c$ for 
$\op \in \{\max, \min, \mathrm{sum}, \avg\}$ counts the satisfaction of traces over a set of locations. 
From \textit{counting} operators, we derive the \emph{everywhere} operator as $\ew \varphi \equiv \ct^{\mathrm{min}} \varphi > 0$, and \emph{somewhere} operator as $\sw \varphi \equiv \ct^{\mathrm{max}} \varphi > 0$. Please refer to \cite{ma2021novel} for the detailed definition and semantics of SaSTL.

\subsection{Interface for Intelligent Assistant}
\label{subsec:interface}

City requirements often have missing or ambiguous information, which may be unnoticed by policy makers. It leads to the demand for human inputs and clarification when converting them into 
formal specifications. Therefore, we design an intelligent assistant interface in CitySpec serving as an intermediary between policy makers and the translation model. It communicates with policy makers and confirms the final requirements through an intelligent conversation interface. 

To briefly describe the communication process, users first input a requirement in English, e.g., ``due to safety concerns, the number of taxis should be less than 10 between 7 am to 8 am''. CitySpec interface passes the requirement to the translation model and gets a formal requirement ($\mathsf{always}_{[7,8]} \mathsf{number~of~taxi}<10$) with the keywords including,  
\begin{itemize}
    \item $\mathsf{entity}$: the requirement's main object, e.g., ``the number'',
    \item $\mathsf{quantifier}$: the scope of an entity, e.g., ``taxi'',
    \item $\mathsf{location}$: the location where this requirement is in effect, which is missing from the above example requirement, 
    \item $\mathsf{time}$: the time period during which this requirement is in effect, e.g., ``between 7 am to 8 am'', 
    \item $\mathsf{condition}$: the specific constraint on the entity, such as an upper or lower bound of $\mathsf{entity}$, e.g., ``10''.  
\end{itemize}

As a result, CitySpec detects that the location information is missing from the user's requirement and generates a query for the user, ``what is the location for this requirement?'' Next, with new information typed in by the user (e.g., ``within 200 meters of all the schools''), CitySpec obtains a complete requirement. 

\sloppy The next challenge is how to confirm the formal specification with policy makers. Since they do not understand the formal equation, we further convert it to a template-based sentence. Therefore, CitySpec presents three formats of this requirements for users to verify, (1) a template-based requirement, e.g., [number] of [taxi] should be [$<$] [10] [between 7:00 to 8:00] [within 200 meters of all the schools], (2) a SaSTL formula $\mathsf{everywhere}_{\mathsf{school}\land[0,200]}\mathsf{always}_{[7,8]} \mathsf{number~of~taxi}<10$, and (3) five key fields detected. Users can confirm or further revise this requirement through the intelligent assistant. 

When policy makers have a large number of requirements to convert, to minimize user labor to input requirements manually, CitySpec also provides the option for them to input requirements through a file. The process is similar to where CitySpec asks users to provide or clarify information until all the requirements are successfully converted through files.


\subsection{Requirement Synthesis}
\label{subsec:requirement}
The amount of city requirement dataset is insufficient to train a decent translation model in an end-to-end manner. As we've discussed in Section \ref{sec:introduction}, it requires extensive domain knowledge in both city and formal specifications and is extremely time-consuming to annotate new requirements. Furthermore, a majority of the existing city requirements are qualitatively or imprecisely written, which cannot be added to the requirement dataset without refinement~\cite{ma2021novel}. 
To mitigate the challenge of small data to build a translation model, we design a novel approach to incorporating city knowledge through controllable requirement synthesis. 
 

There are two main reasons why converting a city requirement to a formal specification is challenging with a small amount of data. First, \textit{the vocabulary of city requirements are very diverse}. For example, requirements from different cities (e.g., Seattle and New York City) or in different domains (e.g., transportation and environment) have totally different vocabulary for entities, locations, and conditions. Second, \textit{the sentence structure (patterns) of requirements vary significantly when written by different people.} It is natural for human beings to describe the same thing using sentences.  

Targeting these two challenges, we first extract city knowledge and build two knowledge datasets, i.e., a vocabulary set and a pattern set. The vocabulary set includes five keys of a requirement, i.e., entity, quantifier, location, time and condition. The pattern set includes requirement sentences with 5 keywords replaced by their labels. For instance, we have a requirement, ``In all \textit{buildings}/$\mathsf{location}$, the average \textit{concentration}/$\mathsf{entity}$ of \textit{TVOC}/$\mathsf{quantifier}$ should be no more than \textit{0.6 mg/m3}/$\mathsf{condition}$ for \textit{every day}/$\mathsf{time}$.'', the pattern extracted is ``In \#$\mathsf{location}$, the average \#$\mathsf{entity}$ of \#$\mathsf{quantifier}$ should be no more than \#$\mathsf{condition}$ for \#$\mathsf{time}$.'' 


We extract the knowledge set from city documents besides requirements so that we are not limited by the rules of requirements and enrich the knowledge of our model.  
For example, we extract 336 patterns and 3061 phrases (530 phrases in $\mathsf{entity}$, 567 phrases in $\mathsf{quantifier}$, 501 phrases in $\mathsf{location}$, 595 phrases in $\mathsf{condition}$, and 868 phrases in $\mathsf{time}$). 





Next, we designed an approach to synthesizing controllable requirement dataset efficiently.
Intuitively, we can go through all the combinations of keywords and patterns to create the dataset of requirements, which is infeasible and may cause the model overfitting to the injected knowledge. 
In order to enhance the model's performance, we need to keep a balance between the coverage of each keyword and the times of keywords being seen in the generation. 
We denote $\lambda$ as the synthesis index, which indicates the \emph{minimum} number of times that a keyword appears in the generated set of requirements. Assuming we have $m$ set of keywords vocabularies $\{V_1, V_2 \dots V_m\}$ and a pattern set as $P$, we have the total number of synthesized requirements $\ell = \lambda \cdot \max(|V_1|, |V_2|, \dots, |V_m|)$. For each set of vocabularies $V_i$, we first create a random permutation of $V_i$ and repeat it until the total number of phrases reaches $\ell$, then we concatenate them to an array $S_i$. Once we obtain $S_1, ... S_m$, we combine them with pattern $P$ to generate a requirement set $R$. Refer to Algorithm \ref{alg:data_enrich} for more details.

\begin{algorithm}[t]
 \caption{Requirement Synthesis}
 \label{alg:data_enrich}
 \footnotesize
\textbf{Input:} $m$ set of keywords vocabularies $\{V_1, V_2 \dots V_m\}$, Pattern $P$, synthesis index $\lambda$ \\
\textbf{Output:} Set of requirements $R$
\begin{algorithmic}
\State Initialize $R$ as an empty set$\{\}$
\State Let $\ell = \lambda \cdot \max(|V_1|, |V_2|, \dots, |V_m|)$
\For{$i \in 1 \dots m$}
\State Initialize $S_i$ as an empty array $S_i = []$
\While{$|S_i| < \ell$}
    \State Create a random permutation of $V_i$: $P = \text{Permutate}(V_i)$ 
    \State Concatenate $P$ to $S_i$: $S_i = \text{Concat}(S_i, P)$
\EndWhile
\EndFor
\For{$j \in 1 \dots \ell$}
    \State Combine keywords $S_1[j], S_2[j] \dots S_m[j]$ with Pattern $P$ to create a requirement $r_j$
    \State Add $r_j$ to the set of requirements $R$
\EndFor
\State \Return $R$
\end{algorithmic}
\end{algorithm}

\subsection{Translation Model}
\label{subsec:translation}

The inputs of the translation model are requirements, and the outputs of this module are formal specifications with token-level classification. 
We implement the translation model with three major components, a learning model, knowledge injection through synthesized requirements, and keyword refinement. 

To be noted, CitySpec does not build its own translation model from scratch. Instead, we tackle the limitation of the traditional language model and improve it for city requirement translation. Therefore, CitySpec is compatible with different language models. 

In this paper, we implement four popular language models, which are Vanilla Seq2Seq, Stanford NLP NER, Bidirectional Long Short Term Memory (Bi-LSTM) + Conditional Random Field (CRF) and Bidirectional Encoder Representations from Transformers (BERT)~\cite{devlin2018bert}. 
We apply our synthesized datasets with different synthesis indexes to inject city knowledge into these language models.  
Then we evaluate the improvement brought by our requirement synthesis approach by testing the performance on real-world city requirements. We present the detailed results and analysis in Section \ref{sec:eval}. 

Additionally, we find that \textit{time, negation and comparison} are the most tricky elements that affect the accuracy of the final specification detection. Therefore, we implement another refinement component in the translation model.   
In general, the $\mathsf{time}$ can be represented in several formats, such as timestamps, or other formats like yyyy-mm-dd and mm-dd-yyyy. To mitigate the confusion that various formats might bring, we apply SUTime~\cite{chang2012sutime} when the $\mathsf{time}$ entity is not given by the translation model. 
pyContextNLP~\cite{chapman2011document} is applied to analyze whether there is a negation in the input sentence. If there is any negation, the comparison symbol is reversed. For instance, if there is a keyword ``greater than'', the comparison symbol is $>$. However, if the whole phrase is ``is not supposed to be greater than'', and a negation is detected, thus the final comparison is $\leq$ instead.

\subsection{\revision{Secured Online Learning}}
\label{subsec:online}
In general, the more clarifications are needed from the users, the worse experience the users will have, especially if users have to clarify the same information repeatedly. For example, if a user from a new city inputs a location that the system fails to detect, the user will be asked to clarify the location information. The user's experience will drop if the system asks him again on the second or third time seeing these words. However, the deep learning-based translation model cannot ``remember'' this information at deployment time. Thus, the first question is that \textit{how can CitySpec learn the new knowledge online? }

Meanwhile, the new information provided by users may also harm the system if it is an incorrect or adversarial example. The second question is that \textit{how can CitySpec validate the new knowledge before learning it permanently?}   

\begin{figure}[t]
    \centering
    \includegraphics[width=0.98\textwidth]{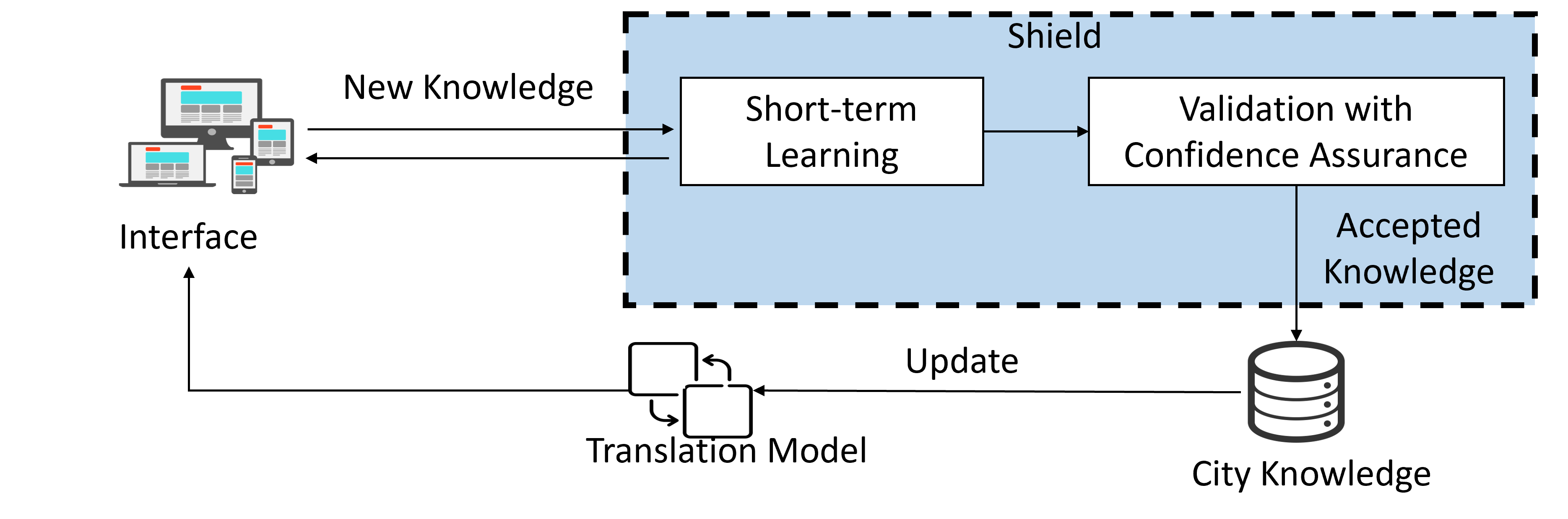}
    \caption{Secured Online Learning}
    \label{fig:online_sample}
    \vspace{-0.8cm}
\end{figure}

\begin{figure}[t]
    \centering
    \includegraphics[width=0.98\textwidth]{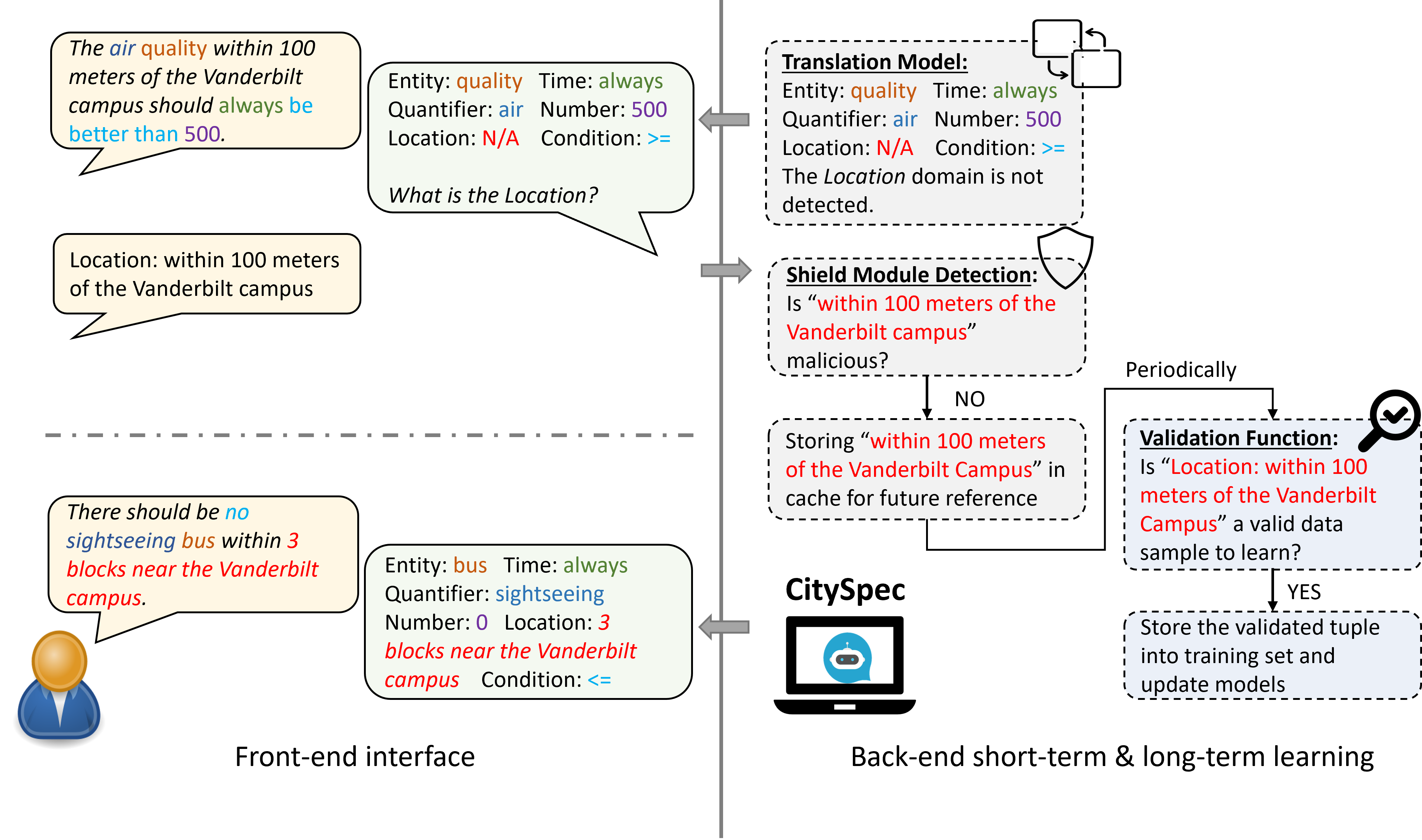}
    \caption{\textcolor{blue}{One Sample Round of Online Learning}}
    \label{fig:online}
    \vspace{-0.8cm}
\end{figure}

Targeting these two research questions, we design an online learning module in CitySpec. As shown in Figure \ref{fig:online}, it has two stages, which are short-term learning and long-term learning. Short-term learning is designed to accommodate the same user in one session of requirement specification with a temporary memory. The question-answer pairs are stored temporarily. When the same occasion occurs, the temporal cache gives instant answers and avoids more user clarifications. 
Long-term learning is designed to adapt the new knowledge to the model permanently after validating its reliability. 
The accepted permanent knowledge is achieved by updating weights via back-propagation on the extended dataset with both initial data and the new input-label pairs stored in the temporary cache. 


\revision{To prevent the injection of malicious or suspicious knowledge into CitySpec, we have implemented two sub-components to secure the online learning session: the Shield Model and the Validation Function. The Shield Model has been designed to segregate non-malicious inputs from malicious ones. As depicted in Fig.\ref{fig:online} and Fig.\ref{fig:online_sample}, the Shield Function protects both short-term and long-term online learning sessions by determining whether the user input is malicious. For instance, when a user provides their location as "within 100 meters of the Vanderbilt campus" in the front-end interface after the Translation Model's output, shown in Fig.\ref{fig:online_sample}, the Shield Model examines it from the backend and prevents any malicious information from being included, as shown on the right side of Fig.\ref{fig:online_sample}. After one user clarification is passed through the Shield Model, it is stored to cache for future references, for example, if the user again inputs some requirement with ``within 100 meters of the Vanderbilt campus'' as the location, CitySpec will directly provide the answer from the cache. Periodically those samples passed the shield function are then passed on to the Validation Function, which further examines whether the user has provided the correct label-phrase pair. As shown in the last few steps in Fig.\ref{fig:online_sample}, the Validation Function verifies whether "within 100 meters of the Vanderbilt campus" should be labeled as a location. Once examined by both sub-components, user inputs will be injected into the city knowledge and used for learning purposes.}

\subsubsection{Validation Function}

In our setting, if CitySpec fails to fulfill all the predefined key domains, it will ask the user for clarification in token-domain pairs. We implement a BERT-based classification model as the validation function and train it on generated city knowledge from all the existing requirements. The inputs of the validation model are the new terms provided by the user, while the outputs are the corresponding keys based on the new terms. To keep CitySpec away from adversarial inputs, we develop a Bayesian CNN-based validation module in CitySpec. The model is to classify the category of a new term with confidence with uncertainty estimation. We apply dropout layers during both training and testing to quantify the model uncertainty~\cite{ma2021predictive}. The inputs of the validation model are the new terms provided by the user, while the outputs are the corresponding keys among those five key elements with an uncertainty level. In brief, a new term-key pair is rejected if (1) the output from the validation function does not align with the given domain key; (2) the validation function has low confidence in the output although it might align with the given domain. In this way, we only accept new city knowledge validated with high confidence.

\subsubsection{Shield Model}

By our design, although the validation function is supposed to serve as an over-watch to prevent CitySpec from any suspicious new knowledge, we still find it vulnerable to adversarial attacks because it directly takes textual inputs from the user. If the user keeps manipulating textual inputs and by enumeration, the user can find adversarial inputs that would confuse the validation function, pass the validation function check, and directly poison our city knowledge. Since the validation function is trained on the city knowledge periodically. If the city knowledge gets poisoned, not only the validation function but also CitySpec will start to malfunction. 

\revision{For example, suppose a truck company desires to increase its revenue despite the existence of current regulations, such as "trucks are not allowed to enter Charlotte Pike from 5 PM to 7 PM on weekdays". In such a scenario, the company may opt to engage in an attack on the system by poisoning the city knowledge, which could involve manipulating the location or time data. By doing so, such regulation would cease to be effectively enforced, enabling the truck company to allocate their trucks onto Charlotte Pike during rush hours. This action, however, poses a significant threat to public safety, given that it could lead to traffic accidents and congestion on the road.}

Thus, we need an additional shield model to guard the validation function so that CitySpec will be hard to attack directly and robust to most adversarial attacks. The shield model, by our design, can detect any malicious behavior when the user clarifies the unfulfilled domain before feeding the clarification to the validation function. Our shield model consists of two separate filters: \textbf{literal correction} and \textbf{inferential mapping}. To test the effectiveness of the shield model, we generate malicious attacks with 12 different kinds of generated adversarial inputs from the most recent works of literature \cite{yoo-qi-2021-towards-improving, garg-ramakrishnan-2020-bae, li-etal-2020-bert-attack, feng-etal-2018-pathologies, pruthi-etal-2019-combating, zang-etal-2020-word, ren-etal-2019-generating, li2018textbugger, jin2020bert, ribeiro-etal-2020-beyond, li-etal-2021-contextualized, gao2018black}.

Those investigated and implemented adversarial attacks \cite{morris2020textattack} aim to mislead the validation function based on existing city knowledge by generating new malicious textual inputs. Although those adversarial attack types differ from each other, we conclude two key steps during those attacks: Word Importance Ranking $S(.)$ and Word Transformation $T(.)$. Given an existing valid input sequence, $X$ and the victim function $F$, in general, an attacker first ranks the sub-tokens and gets the most `important' ones $x_{i,j,k,..} \in X$ based on selected metrics. Then the attacker will transform those selected tokens into other textual inputs $x^{'}_{i,j,k,...}$ and replace the original ones in $X$. The newly assembled sequence $X^{'}$ is passed to $F$. If $F$ changes its output, then the attack is considered successful. Thus, a successful attack would satisfy $F(T(S(X))) \neq F(X)$. Some of those adversarial attacks directly skip the word importance ranking process, even though the rest of those do rank tokens, different assumptions are made and different ranking metrics are applied. We choose to let the shield model focus more on the word transformation during detection. In those attacks, there are two main genres of word transformation: inner-word transformation and inter-word transformation. Inner-word transformations change the characters within selected tokens, e.g., change ``energy consumption'' to ``energy consumptlon''. Inter-word transformations change the whole word instead based on some replacement strategy, e.g., change ``available space'' to ``free space''. More details can be found in Table. \ref{tab:attacks}.

We develop a \textbf{literal correction} filter to detect and leave out any literal disinformation provided in the input tokens. The literal correction is designed to mainly focus on the inner-word transformation. This filter consists of two main components: a language model and a type checker. The language model is trained on over 1,500 existing city requirements. Given a seed token, the language model generates the most `reasonable' sequence by maximum likelihood estimation. Based on the reference word provided by the language model, the type checker searches for all possible variant words based on the reference word without changing too much in spelling. The input sequence is considered malicious if any correction is involved in the initial inputs. E.g., the user types in ``in the m0rninGs''. If the language model gives ``in the mornings'' as a reference, then the type checker will try to recover ``m0rninGs'' to ``mornings''. If the edit distance is allowed, then the input sequence will be corrected to ``in the mornings''. As observed, the corrected sequence is no longer the same as it was first typed in, then this sequence is considered malicious. This Literal Correction filter is considered as hard for direct attacks because the attacker has no access to (1) the existing 1,500 city requirements; (2) the details in the language model; or (3) the edit distance budget in the type checker.

However, this Literal Correction filter is built upon a fixed vocabulary and mainly focuses on word spelling and phrase composing. It ignores the inferential understanding of a given phrase. Thus, we need another filter that suffers less from the limited vocabulary and has more prior knowledge in natural language inferential understanding.

We leverage the prior knowledge brought by the BERT model and develop an \textbf{inferential mapping} filer. It first embeds input sequences with arbitrary lengths into vectors with a fixed length \cite{wang2021phrase}. This embedding process can also be understood as mapping textual sequences into a numeric hyperspace. Based on studies in \cite{wang2021phrase}, this hyperspace is declared to manage to put phrases with similar inferential information closer. E.g., the euclidean distance between ``at the gates'' and ``at the doors'' after embedding is 8.805, however, the distance between ``at the gates'' and ``on the campus'' is 14.95. Based on this characteristic, this inferential mapping filter can mitigate the issues brought by inter-word transformations like word insertion. After mapping all phrases into the hyperspace as vectors with a fixed length, we employ multi-layer perceptrons to draw the decision boundary in the hyperspace. To make this inferential mapping filter more robust to direct attacks, we even mask the embedding vectors with trainable weights before passing them forward to the downstream perceptrons. In conclusion, this filter is considered difficult to attack because (1) all textual information is encrypted by masked phrase embeddings before feeding to the discriminator, which means there is no strong direct relationship between the textual inputs and their corresponding numeric representations; (2) the attacker does not have the access to the model details of the downstream perceptrons.

\begin{sidewaystable}
\caption{\textcolor{blue}{Adversarial Attacks Overview}}
\footnotesize
\centering
{%
\color{blue}\begin{tabular}
{|M{1in}|M{1.1in}|M{1.1in}|M{2.3in}|M{2.3in}|}
\hline
\textbf{Attack Types}   & \textbf{Word Importance Ranking} & \textbf{Word Transformation Strategy}      & \textbf{Original Tokens}               & \textbf{Perturbed Tokens}            \\ \hline\hline
A2T            & Gradient wrt. Loss Function          & Iterative Synonym Replacement               & For any source of sound ..., shall not exceed the \textbf{peak levels} of ...                  & For any source of sound ..., shall not exceed the \textbf{peak tiers} of ...                      \\ \hline
BAE            & N/A                     & Bert Generation                   & No person shall \textbf{use, operate, or maintain} an alert system at ...    & No person shall \textbf{use, access, or maintain} an alert system at ...     \\ \hline
BertAttack     & Forward Result Changes after Permutation        & Bert Generation                   & No operator of a sidewalk cafe shall be assigned all the \textbf{available space} within ...              & No operator of a sidewalk cafe shall be assigned all the \textbf{available space} within ...                       \\ \hline
InputReduction & Confidence Change from Prediction Dist.       & Repetitive Word Removal           & DC minimum roof reflectance: three year-aged \textbf{solar reflectance} of 0.55            & DC minimum roof reflectance: three year-aged \textbf{reflectance} of 0.55                           \\ \hline
Pruthi's       & N/A                     & Character-level Perturbation      & All Refrigerators and ... shall have maximum \textbf{energy consumption} less than ...           & All Refrigerators and ... shall have maximum \textbf{energy consumptlon} less than ...             \\ \hline
PSO            & N/A                     & Word Replacement based on Sememes & The \textbf{location} of a vending machine shall ...                    & The \textbf{locality} of a vending machine shall ...                     \\ \hline
PWWS           & Word Saliency Change on Output Dist.          & Synonym Replacement               & \textbf{Construction and demolition} within one thousand (1,000) feet of a residential property is prohibited when ...  & \textbf{Building and demolition} within one thousand (1,000) feet of a residential property is prohibited when ...       \\ \hline
TextBugger     & Confidence Change from Prediction Dist.        & Character-level Perturbation     & ... the \textbf{material composition} of the pavement system layers ...         & ... the \textbf{material composing} of the pavement system layers ...           \\ \hline
TextFooler     & Forward Result Changes after Deletion          & Synonym Replacement               & No \textbf{idling vehicles} are allowed about or on any ... schools           & No \textbf{unused vehicles} are allowed about or on any ... schools               \\ \hline
DeepWordBug    & LSTM-based Score        & Character-level Perturbation      & ... no person shall \textbf{stop , park , or leave standing} any vehicle ... & ... no person shall \textbf{stCop, pakr, or leav standing} any vehicle\\ \hline
CheckList      & N/A                     & N/A                               & N/A                           & N/A                           \\ \hline
CLARE          & N/A                     & Replace, Insert and Merge         & ... the total gross \textbf{floor area} of all structures ...                   & ... the total gross \textbf{floor residential area} of all structures ...       \\ \hline
\end{tabular}
}
\label{tab:attacks}
\end{sidewaystable}

\section{Evaluation }
\label{sec:eval}

In this section, we evaluate our CitySpec system from five aspects, including (1) comparing different language models on the initial dataset without synthesizing, (2) analyzing the effectiveness of the synthesized requirements by enhancing the models with city knowledge, (3) evaluating the performance of the online validation model, (4) testing CitySpec's adaptability in different cities and application domains, and (5) an overall case study. We use the city requirement {dataset} described in Section \ref{sec:motivation}. 
To evaluate the prediction of keywords and mitigate the influence caused by different lengths requirements, 
we choose to use \textit{token-level accuracy (token-acc)} and \textit{sentence-level accuracy (sent-acc)} as our main {metrics}. The token-level accuracy aims to count the number of key tokens that are correctly predicted. The sentence-level accuracy counts the prediction as correct only when the whole requirement is correctly translated to a formal specification using SaSTL. Thus, sentence-level accuracy serves as a very strict criterion to evaluate the model performance. 
We also provide the results using other common metrics including precision, recall, and F-1 score. 
The experiments were run on a machine with 2.50GHz CPU,
32GB memory, and Nvidia GeForce RTX 3080Ti GPU.

\subsection{Performance of language models on the initial dataset}

\begin{table*}[t]
    \centering
        \caption{Performance of language models on the initial dataset}
        \scriptsize
\resizebox{\textwidth}{!}{%
 \begin{tabular}{||c | c | c| c | c | c ||} 
 \hline
 \textbf{Model} & \textbf{Token-Acc} & \textbf{Sent-Acc} & \textbf{F-1 Score} & \textbf{Precision} & \textbf{Recall} \\ [0.2ex] 
 \hline\hline
 \textbf{Vanilla Seq2Seq} & \ 10.91 $\pm$ 0.57 \% & \ 1.38 $\pm$ 0.49\% & \ 24.12 $\pm$ 0.24 \% & \ 65.58 $\pm$ 7.95 \% & \ 14.81 $\pm$ 1.58 \%\\ 
 \hline
\textbf{BiLSTM + CRF} & \ 77.59 $\pm$ 0.52 \% & \ 60.82 $\pm$ 1.22 \% & \ 80.46 $\pm$ 0.84 \% & \ 81.11 $\pm$ 1.38 \% & \ 79.83 $\pm$ 7.24 \%\\
 \hline
 \textbf{BERT} & \ 80.41 $\pm$ 0.07 \% & \ 59.02 $\pm$ 0.42 \% & \ 81.43 $\pm$ 0.01 \% & \ 78.62 $\pm$ 0.01 \% & \ 84.46 $\pm$ 0.01 \%\\
 \hline
\end{tabular}}
    \label{tab:initial}
    \vspace{-0.3cm}
\end{table*}

As a baseline of translation model without city knowledge,  
we first evaluate the performance of CitySpec using different language models, including Vanilla Seq2Seq, pretrained Stanford NER Tagger, Bi-LSTM + CRF, and BERT on the initial dataset. We present the results in Table \ref{tab:initial}. 

We make the following observations from the results. 
First, the overlap between Stanford Pretrained NER Tagger prediction and vocabulary is only 9 out of 729. The pretrained tagger tends to give locations in higher granularity. Since this task is a city-level, more detailed location information is stated in a lower granularity by providing the street name, building name, or community name. The $\mathsf{location}$ domain in the pretrained tagger gives more high-level information like city name, state name, or country name. For example, ``34th Ave in Nashville, the state of Tennessee'' is annotated as $\mathsf{location}$ in this task, however, the pretrained NER tagger gives ``Tennessee" as $\mathsf{location}$ instead. 

Secondly, the testing token-acc from Vanilla Seq2Seq is 10.91\% on average. Other metrics also indicate that Vanilla Seq2Seq has trouble recognizing the patterns in sequential keyword labeling. 
The Vanilla Seq2Seq model suffers from data scarcity and has difficulty recognizing the general patterns in the training samples due to the small size of the dataset.

Thirdly, the Bi-LSTM + CRF and BERT model achieve better performance than other models, and BERT models often outperform other models with lower standard deviation. 
However, the highest token-level accuracy achieved is 80.41\%, which is still not high enough for an accuracy-prioritized task. A different key may change the requirement entirely. For example, the ``width'' of ``car windshield'' and the ``width'' of ``car'' focus on completely different aspects, although the keywords ``car windshield'' and ``car'' have an only one-word discrepancy. Meanwhile, the best sentence-level accuracy achieved is 60.82\%, which means that about 40\% of the requirements are falsely translated. Assuming the policy makers fix these requirements through the intelligent assistant interface, it is time-consuming and reduces the user experience. Even worse, it may bring safety issues to the monitoring system without noticing.




{In summary, the results indicate that existing language models are not sufficient to serve as the translation model for CitySpec directly. There is a high demand for injecting city knowledge to build the translation model. 
}

\subsection{Requirement Synthesis with City 
Knowledge}

\begin{figure}[t]
    \centering
    \includegraphics[width=0.98\textwidth]{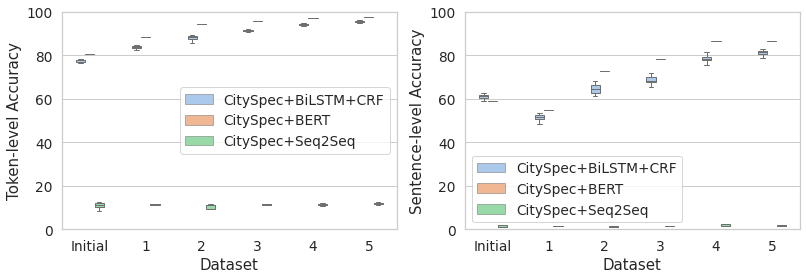}
    \caption{Performance improvement brought by requirement synthesis}
    \label{fig:exp_res}
    \vspace{-0.6cm}
\end{figure}

\begin{figure}[t]
    \centering
    \includegraphics[width=0.98\textwidth]{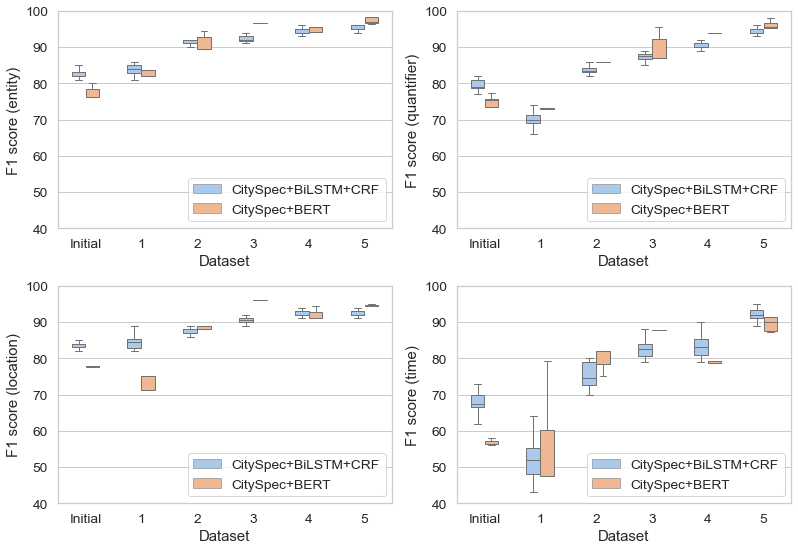}
    \caption{F-1 scores on four keywords}
    \label{fig:key_res}
    \vspace{-0.5cm}
\end{figure}

Next, we evaluate CitySpec's performance with our controllable synthesized requirements. For a fair comparison, we do not use the requirements in the testing set to synthesize requirements. We ensure that the trained model has not seen the requirements in the testing set in either the knowledge injection or training phases.  
We apply different synthesis indexes to test the effects on the prediction performance. We present the overall results on token-level and sentence-level accuracy in Figure \ref{fig:exp_res}, and F-1 scores on individual keyword in Figure \ref{fig:key_res}. In the figures, x-axis represents the synthesis index. When the index equals to ``inital'', it shows the results without synthesis data.

From the results, we find that, for BERT and Bi-LSTM, there is an overall increase in performance in all token-level accuracy, sentence-level accuracy, overall F-1 score, and F-1 score on keywords. For example, BiLSTM+CRF's token-level accuracy increases from 77.59\% to 97\% and sentence-level accuracy increases from 60.82\% to 81.3\%, BERT's sentence-level accuracy increases from 59.02\% to 86.64\%. 

\textit{In summary, the results show that injecting city knowledge with synthesized requirements boosts the translation model significantly. While improving policy maker's user experience with higher accuracy and less clarifications, it also enhances the safety of the monitoring system potentially.}


\subsection{Performance on Online Validation }


We evaluate the validation model through simulating four different testing scenarios: (I) randomly generated malicious input based on the permutation of letters and symbols; (II) all street names in Nashville; (III) real city vocabulary generated from Nashville requirements; (IV) generated float numbers with different units. 

First of all, the accuracy of validation model is very high. When the uncertainty threshold is set to 0.5, i.e., all inputs cause an uncertainty higher than 0.5 will be ruled out, CitySpec gives 100\% success rate against scenario I among 2,000 malicious inputs, 91.40\% acceptance rate among 2,107 samples in scenario II, 92.12\% acceptance rate among 596 samples in scenario III, and 94.51\% acceptance rate among 2,040 samples in scenario IV. 
Additionally, we find that the validation function easily confuses $\mathsf{entity}$ with $\mathsf{quantifier}$ if no further guidance is offered. We look into dataset and figure out $\mathsf{entity}$ and $\mathsf{quantifier}$ are confusing to even humans without any context information. Take the requirement ``In all buildings, the average concentration of Sulfur dioxide (SO2) should be no more than 0.15 mg/m3 for every day.'' As an example, $\mathsf{entity}$ is ``concentration'' and $\mathsf{quantifier}$ is ``Sulfur dioxide (SO2)''. If the requirement is changed to ``The maximum level of the concentration of Sulfur dioxide (SO2) should be no more than 0.15 mg / m3 for every day.'', then $\mathsf{entity}$ is ``maximum level'' and $\mathsf{quantifier}$ is ``concentration'' instead. In addition, terms like ``occupancy of a shopping mall'', ``noise level at a shopping mall'', and ``the shopping mall of the commercial district'' also introduce confusion between $\mathsf{location}$, $\mathsf{entity}$ and $\mathsf{quantifier}$, since the same token ``shopping mall'' can be $\mathsf{entity}$, $\mathsf{quantifier}$ or $\mathsf{location}$ in certain cases. 

\textit{The results show that the validation algorithm can effectively accept new city knowledge, prevent adversarial inputs and safeguard online learning. Therefore, CitySpec reduces unnecessary interactions between policy makers and the system and increases efficiency.}



\subsection{Adaptability to different scenarios}


\begin{table*}[t]
\caption{Adaptability on different cities in terms of token-level accuracy, sent-level accuracy, and overall F-1 score}
\scriptsize
\centering
\begin{tabular}{|c|ccc|ccc|}
\hline
\textbf{City}                       & \multicolumn{3}{c|}{\textbf{Seattle}}                                           & \multicolumn{3}{c|}{\textbf{Changsha}}                                          \\ \hline\hline
\textbf{Metrics}                    & \multicolumn{1}{c|}{\textbf{ToeknAcc}} & \multicolumn{1}{c|}{\textbf{SentAcc}} & \textbf{F-1}     & \multicolumn{1}{c|}{\textbf{ToeknAcc}} & \multicolumn{1}{c|}{\textbf{SentAcc}} & F-1     \\ \hline
\begin{tabular}[c]{@{}c@{}}\textbf{Non-adaptive w/} \\ \textbf{BiLSTM+CRF}\end{tabular} & \multicolumn{1}{c|}{84.91\%}  & \multicolumn{1}{c|}{48.00\%} & 77.60\% & \multicolumn{1}{c|}{86.61\%}  & \multicolumn{1}{c|}{61.20\%} & 84.10\% \\\hline
\begin{tabular}[c]{@{}c@{}}\textbf{Adaptive w/} \\ \textbf{BiLSTM+CRF}\end{tabular}      & \multicolumn{1}{c|}{96.05\%}  & \multicolumn{1}{c|}{84.80\%} & 93.75\% & \multicolumn{1}{c|}{95.27\%}  & \multicolumn{1}{c|}{83.20\%} & 93.57\% \\ \hline
\begin{tabular}[c]{@{}c@{}}\textbf{Non-adaptive w/} \\ \textbf{BERT}\end{tabular}       & \multicolumn{1}{c|}{80.38\%}  & \multicolumn{1}{c|}{46.40\%} & 76.80\% & \multicolumn{1}{c|}{86.10\%}  & \multicolumn{1}{c|}{58.00\%} & 83.87\% \\\hline
\begin{tabular}[c]{@{}c@{}}\textbf{Adaptive w/} \\ \textbf{BERT}\end{tabular}           & \multicolumn{1}{c|}{95.10\%}  & \multicolumn{1}{c|}{80.70\%} & 90.28\% & \multicolumn{1}{c|}{97.16\%}  & \multicolumn{1}{c|}{88.40\%} & 96.88\% \\ \hline\hline
\textbf{City}                       & \multicolumn{3}{c|}{\textbf{Charlottesville}}                                   & \multicolumn{3}{c|}{\textbf{Jacksonville}}                                      \\ \hline\hline
\textbf{Metrics}                    & \multicolumn{1}{c|}{\textbf{ToeknAcc}} & \multicolumn{1}{c|}{\textbf{SentAcc}} & \textbf{F-1}     & \multicolumn{1}{c|}{\textbf{ToeknAcc}} & \multicolumn{1}{c|}{\textbf{SentAcc}} & F-1     \\ \hline
\begin{tabular}[c]{@{}c@{}}\textbf{Non-adaptive w/} \\ \textbf{BiLSTM+CRF}\end{tabular} & \multicolumn{1}{c|}{90.00\%}  & \multicolumn{1}{c|}{65.62\%} & 86.48\% & \multicolumn{1}{c|}{77.32\%}  & \multicolumn{1}{c|}{35.20\%} & 81.54\% \\\hline
\begin{tabular}[c]{@{}c@{}}\textbf{Adaptive w/} \\ \textbf{BiLSTM+CRF}\end{tabular}     & \multicolumn{1}{c|}{96.82\%}  & \multicolumn{1}{c|}{89.29\%} & 95.40\% & \multicolumn{1}{c|}{97.35\%}  & \multicolumn{1}{c|}{88.80\%} & 96.02\% \\ \hline
\begin{tabular}[c]{@{}c@{}}\textbf{Non-adaptive w/} \\ \textbf{BERT}\end{tabular}       & \multicolumn{1}{c|}{93.44\%}  & \multicolumn{1}{c|}{73.21\%} & 90.46\% & \multicolumn{1}{c|}{90.21\%}  & \multicolumn{1}{c|}{56.40\%} & 81.60\% \\\hline
\begin{tabular}[c]{@{}c@{}}\textbf{Adaptive w/} \\ \textbf{BERT}\end{tabular}            & \multicolumn{1}{c|}{97.53\%}  & \multicolumn{1}{c|}{87.05\%} & 94.31\% & \multicolumn{1}{c|}{96.27\%}  & \multicolumn{1}{c|}{83.20\%} & 92.59\% \\ \hline
\end{tabular}

\label{tab:cities}
\end{table*}

\begin{table*}[t]
\caption{Adaptability on different topics in terms of token-level accuracy, sent-level accuracy, and overall F-1 score}
\scriptsize
\begin{tabular}{|c|ccc|ccc|}
\hline
\textbf{Topic}                      & \multicolumn{3}{c|}{\textbf{Noise Control}}                                                   & \multicolumn{3}{c|}{\textbf{Public Access}}                                                   \\ \hline\hline
\textbf{Metrics}                    & \multicolumn{1}{c|}{\textbf{ToeknAcc}} & \multicolumn{1}{c|}{\textbf{SentAcc}} & \textbf{F-1} & \multicolumn{1}{c|}{\textbf{ToeknAcc}} & \multicolumn{1}{c|}{\textbf{SentAcc}} & \textbf{F-1} \\ \hline
\begin{tabular}[c]{@{}c@{}}\textbf{Non-adaptive w/} \\ \textbf{BiLSTM+CRF}\end{tabular} & \multicolumn{1}{c|}{77.82\%}           & \multicolumn{1}{c|}{41.56\%}          & 77.83\%      & \multicolumn{1}{c|}{73.99\%}           & \multicolumn{1}{c|}{44.07\%}          & 74.41\%      \\\hline
\begin{tabular}[c]{@{}c@{}}\textbf{Adaptive w/} \\ \textbf{BiLSTM+CRF}\end{tabular}     & \multicolumn{1}{c|}{95.82\%}           & \multicolumn{1}{c|}{90.68\%}          & 94.46\%      & \multicolumn{1}{c|}{97.68\%}           & \multicolumn{1}{c|}{74.80\%}          & 97.39\%      \\ \hline
\begin{tabular}[c]{@{}c@{}}\textbf{Non-adaptive w/} \\ \textbf{BERT}\end{tabular}        & \multicolumn{1}{c|}{84.15\%}           & \multicolumn{1}{c|}{58.62\%}          & 81.05\%      & \multicolumn{1}{c|}{83.59\%}           & \multicolumn{1}{c|}{54.17\%}          & 78.93\%      \\\hline
\begin{tabular}[c]{@{}c@{}}\textbf{Adaptive w/} \\ \textbf{BERT}\end{tabular}           & \multicolumn{1}{c|}{98.07\%}           & \multicolumn{1}{c|}{88.31\%}          & 92.98\%      & \multicolumn{1}{c|}{97.43\%}           & \multicolumn{1}{c|}{88.75\%}          & 95.75\%      \\ \hline\hline
\textbf{Topic}                      & \multicolumn{3}{c|}{\textbf{Indoor Air Control}}                                              & \multicolumn{3}{c|}{\textbf{Security}}                                                        \\ \hline\hline
\textbf{Metrics}                    & \multicolumn{1}{c|}{\textbf{ToeknAcc}} & \multicolumn{1}{c|}{\textbf{SentAcc}} & \textbf{F-1} & \multicolumn{1}{c|}{\textbf{ToeknAcc}} & \multicolumn{1}{c|}{\textbf{SentAcc}} & \textbf{F-1} \\ \hline
\begin{tabular}[c]{@{}c@{}}\textbf{Non-adaptive w/} \\ \textbf{BiLSTM+CRF}\end{tabular} & \multicolumn{1}{c|}{81.51\%}           & \multicolumn{1}{c|}{46.80\%}          & 76.22\%      & \multicolumn{1}{c|}{72.11\%}           & \multicolumn{1}{c|}{28.80\%}          & 62.93\%      \\\hline
\begin{tabular}[c]{@{}c@{}}\textbf{Adaptive w/} \\ \textbf{BiLSTM+CRF}\end{tabular}     & \multicolumn{1}{c|}{95.58\%}           & \multicolumn{1}{c|}{80.00\%}          & 87.68\%      & \multicolumn{1}{c|}{94.39\%}           & \multicolumn{1}{c|}{94.37\%}          & 92.34\%      \\ \hline
\begin{tabular}[c]{@{}c@{}}\textbf{Non-adaptive w/} \\ \textbf{BERT}\end{tabular}       & \multicolumn{1}{c|}{78.51\%}           & \multicolumn{1}{c|}{31.20\%}          & 73.88\%      & \multicolumn{1}{c|}{79.31\%}           & \multicolumn{1}{c|}{45.60\%}          & 77.50\%      \\\hline
\begin{tabular}[c]{@{}c@{}}\textbf{Adaptive w/} \\ \textbf{BERT}\end{tabular}           & \multicolumn{1}{c|}{95.31\%}           & \multicolumn{1}{c|}{74.40\%}          & 93.60\%      & \multicolumn{1}{c|}{95.41\%}           & \multicolumn{1}{c|}{82.80\%}          & 93.95\%      \\ \hline
\end{tabular}
\label{tab:topics}
\end{table*}

In this section, we analyze CitySpec's adaptability in different cities and different domains.  
Different cities have different regulation focuses and their city-specific vocabulary. For example, in the city of Nashville, $\mathsf{location}$ names like ``Music Row'', ``Grand Ole Opry'' will probably never appear in any other cities. We select four cities, Seattle, Charlottesville, Jacksonville, and Changsha, with different sizes and from different countries as case studies. We separate the requirements of each mentioned city and extract the city-wise vocabulary based on each city independently. Each of four constructed pairs consists of: vocabulary I, which is extracted from the requirements from one city only, and vocabulary II, which is extracted from the requirements from all the cities but that specific one city. Injected knowledge is measured using the number along with the ratio of how much of the unique vocabulary one city causes. We augment vocabulary II using 5 as the synthesis index and train a model on vocabulary II. As a result, the trained model is isolated from the vocabulary information from that one specific city. Afterward, we test the trained model performance on the generated requirements using vocabulary I. We pick CitySpec with Bi-LSTM + CRF and CitySpec with BERT in this scenario. We employ the validation function to validate all vocab in vocabulary I and pass the validated ones to vocabulary II. After that, we have a validated vocabulary including vocabulary II and validated vocabulary I. The deployed model is fine-tuned based on the validated vocabulary using few-shot learning.

From the results shown in Table \ref{tab:cities}, we observe that (1) although CitySpec immigrates  to a completely unknown city, it is still able to provide satisfying performance, e.g., 84.9\% token-acc and 77.6\% F-1 score in Seattle, but the sent-acc tends to be low. (2) With new knowledge injected, the performance increases significantly, e.g., Sent-Acc for Seattle increases from 48\% to 84.8\% with BiLSTM+CRF, and from 46.4\% to 80.7\% with BERT.  

\revision{We further study the practical impacts after having such sent-acc and token-acc. From its definition, when a sent-acc of 84.80\% is achieved, which is the adapted performance of BiLSTM + CRF from the Seattle experiment, it reflects that among all requirements, 84.80\% of them are predicted entirely correctly for both in-domain and out-domain tokens. For example, given ``For all the zones in Civil Building Engineering Group II , retail sale should be less than 10 dB(A) at any time.'' as input, the translation model produces output that can be understood as domain-keyword tuples. In the given requirement, there are words marked as keywords that belong to a predefined category, for example, ``10 dB(A)'' is $\mathsf{number}$. There are also some other words do not belong to any category, for example, ``should be''. To add this requirement to the set of correct predictions using sent-acc, the model needs to predict all tokens into the correct category for both words, no matter whether they belong to any category or not. However, token-acc performs differently from sent-acc since it counts the ratio of correct prediction of all the tokens. Thus, sent-acc is a stricter metric compared to token-acc. }

\revision{We study the performance of the translation model by checking its incorrect predictions. We conclude there are two main situations when the translation model makes mistakes so that the sent-acc would be affected. Situation I: nuance differences that will not affect the overall understanding. Situation II: significant differences caused by the translation model's failure to understand the requirement correctly. Take the requirement ``The sound level in the residential district should always be less than 1,000 feet.'' for example, if the model gives ``$\mathsf{quantifier}$: residential district'' as one of the predictions while the true label is ``$\mathsf{quantifier}$: the residential district'', then this prediction sequence will not be considered correct in terms of sent-acc although this kind of nuance difference does not affect the overall understanding. In another situation where the model gives ``$\mathsf{quantifier}$: district'' instead of ``$\mathsf{number}$: the residential district'', the model predictions should not be adopted. To mitigate the problems brought by these two situations, we rely on the user to validate and modify the prediction of the model if necessary, interactively during the conversation (the results will be applied if it goes through the shield model check). There is also a window on our interface that reports a  summary of the generated specification but in a more readable format \cite{9820752}.}

We also explore CitySpec's adaptability to different topics. We choose four topics including noise control, indoor air control, security, and public access. The results also show that (1) even though CitySpec has not seen vocabulary from a totally different topic, it still gives a competitive performance; (2) online learning brings obvious improvements when adaptation is further applied.

\textit{In summary, it indicates the capability of CitySpec in both city and domain adaptation. It can also adapt to new requirements evolving overtime. 
Moreover, with a different set of domain-specified knowledge, CitySpec can be potentially applied to other application domains (e.g., healthcare). }

\subsection{Performance of Shield Model}

\newcontent{We first launch 12 different adversarial attacks targeting the validation function without any protection from the shield. Only a limited amount of city knowledge is exposed to these attackers for adversarial sample generation. For those attacks that are not in a black-box fashion, we allow them to query the validation function arbitrarily. Because in practical usage, we assume the partial city knowledge has been exposed to the user so that the user could understand and compose city requirements based on that. And since the conversation assistance system is developed in a slot-filling fashion, it will continue for good if not all required slots are fulfilled by the user, so  the model query budget is set to infinity in this case. After these attacks, we employ the shield model to defend against those adversarial samples with three different settings: (1) Literal Correction layer alone; (2) Inferential Mapping layer alone; (3) Inferential Mapping layer with Literal Correction layer as a pre-check. The defense rate is measured as the number of filtered-out adversarial samples divided by the number of remaining adversarial samples. }

\newcontent{As we can tell from the first column in Table \ref{tab:shield}, the validation function is considered as vulnerable to 10 out of those 12 attacks without protection from the shield model, for instance, BertAttack yields 94.00\% as the attack success rate. Thus there is an urgent demand for a reliable protection mechanism deployed against these adversarial attacks. In setting one, although the Literal Correction layer alone gives around 63.11\% defense success rate on average over 12 adversarial attacks, we still find there are several shortcomings in the Literal Correction layer. After error analysis, the Literal Correction filter does not work as expected when (1) the language model fails to give any reference based on the perturbed word, which means the perturbed word never appeared in the existing requirements; (2) the type checker fails to correct the word in a certain amount of edits. E.g., the user inputs ``in the tr1anGl3s''. If the language model fails to give ``triangles'' as a reference, which means the word ``triangle'' never appears in the requirements, then the type checker will no longer work. Even if the language model generates ``in the triangles'' as a reference when the edit distance budge is set to 2, then the type checker will fail to convert ``tr1anGl3s'' to ``triangles''.}

\newcontent{In setting two, although the Inferential Mapping layer yields around 93.09\% as a defense rate, we still find it having trouble gaining a low false-positive rate when faced with non-malicious samples. We pass all legal input samples in our city knowledge to the Inferential Mapping layer and find it reports 31.32\% as a false-positive rate when faced with phrases categorized in $\mathsf{entity}$. We want the shield model to filter out as many malicious samples as possible while as few legal samples as possible. After error analysis, we find the Inferential Mapping layer ignores typos in phrases due to its focus on semantic understanding. For example, if the malicious input is ``in the morinigns'' and there is ``in the mornings'' marked as non-malicious, the Inferential Mapping layer will be confused since the euclidean distance between these two phrases is only 6.21. In other words, if there is only a small typo in input samples, the Inferential Mapping layer will ignore it and treat them similarly to ones without typos. In consequence, the false-positive rate is increased.}

\newcontent{In setting three, the Literal Correction layer helps the Inferential Mapping layer filter out typos first. At the same time, the Inferential Mapping layer leverages its advantages in natural language understanding to identify samples that the Literal Correction layer fails to handle. The overall defense rate is reported to be around 97.78\% and even yields a 100\% defense rate when faced with adversarial attacks like A2T, Pruthi's algorithm, and DeepWordBug. After testing this combination of layers on legal inputs, the false-positive rate is 3.09\% on average over the five domains.}

\newcontent{\textit{In summary, the shield model is necessary for CitySpec because it helps the validation function guard the city knowledge against being poisoned. The combination of both layers as the shield model is effective when faced with 12 adversarial attacks without overkilling non-malicious samples.}}

\subsection{Emulation}
Due to the absence of a real city policy maker, we emulate the process of using CitySpec by taking the real-world city requirements and assuming that they are input by policy makers. Specifically, this case study shows the iteration of communication between CitySpec and the policy maker to clarify the requirements. 
We emulate this process 20 times with 100 requirements randomly selected from our datasets each time. The results show that the average and maximum rounds of clarification are 0.8 and 4 per requirement, respectively, due to the missing or ambiguous information. Averagely, 28.35\% of requirements require clarification on location. 
For example, ``No vendor should vend after midnight.'', CitySpec asks users to clarify the time range for ``after midnight'' and the location defined for this requirement. 
Overall, CitySpec obtains an average sentence-level accuracy of 90.60\% (with BERT and synthesize index = 5). The case study further proves the effectiveness of CitySpec in city requirement specification.

\begin{table}[]
\caption{\textcolor{black}{Shield Defense Rate Against Adversarial Attacks (SR: Success Rate)}}
\scriptsize
\centering
{%
\changetable{
\begin{tabular}{|c|c|c|c|c|}
\hline
Attack Types   & Attack SR & \begin{tabular}[c]{@{}c@{}}Attack SR after \\ \textbf{Literal Correction}\end{tabular} & \begin{tabular}[c]{@{}c@{}}Attack SR after \\  \textbf{Inferential Mapping}\end{tabular} & \begin{tabular}[c]{@{}c@{}}Attack SR after \\  \textbf{both layers}\end{tabular} \\ \hline
A2T            & 9.11\%              & 3.12\%              & 0.72\%              & 0\%                                \\ \hline
BAE            & 8.63\%              & 3.11\%              & 1.52\%              & 0.25\%                              \\ \hline
BertAttack     & 94.00\%             & 61.39\%             & 7.91\%              & 4.31\%                              \\ \hline
InputReduction & 0\%                 & N/A                 & N/A                 & N/A                                  \\ \hline
Pruthi's       & 85.85\%             & 0.96\%              & 0.96\%              & 0\%                                \\ \hline
PSO            & 69.78\%             & 30.63\%             & 3.36\%              & 1.92\%                              \\ \hline
PWWS           & 73.14\%             & 34.05\%             & 3.59\%              & 1.92\%                              \\ \hline
TextBugger     & 21.58\%             & 11.12\%             & 0.48\%              & 0.48\%                              \\ \hline
TextFooler     & 90.17\%             & 39.81\%             & 5.04\%              & 2.39\%                              \\ \hline
DeepWordBug    & 82.73\%             & 1.43\%              & 1.19\%              & 0\%                                \\ \hline
CheckList      & 0\%                 & N/A                 & N/A                 & N/A                                  \\ \hline
CLARE          & 43.14\%             & 18.50\%             & 6.91\%              & 2.38\%                              \\ \hline
\end{tabular}
}
}
\label{tab:shield}
\end{table}

\section{Case Study} 

\newcontent{To demonstrate the effectiveness of CitySpec, we also launch a real user case study on 18 participants with different backgrounds: 7 participants with a Computer Science-related background, 3 with Education-related background, and 2 with Finance-related background. In addition, We have 1 participant for each domain among Biology, Environmental Engineering, Maths, Law, and Art. The goal of this real user case study targets to test the performance of CitySpec (1) while helping the user complete requirements in smart city scenarios; (2) while helping the user complete requirements in an unseen domain; (3) in telling malicious inputs while conveying service; (4) while handling unseen knowledge with the help from its online learning features.}


\subsection{CitySpec in Smart City Scenarios}
\newcontent{To test the usability and adaptability of CitySpec in smart city, We ask the participants to read and type in 8 randomly selected requirements sequentially to CitySpec, then record the time and the number of interactions needed to complete the requirement specification along with subjective user scores. We find the number of interactions is highly related to the input requirement itself, so instead of directly using the number of interactions needed to measure the usability of CitySpec, we, as authors of this work, provide a reference number of interactions of each requirement. We take the difference between the number of user interactions and the number of reference interactions as a measurement of the usability of CitySpec.}

\begin{figure}[t]
    \centering
    \includegraphics[width=\textwidth]{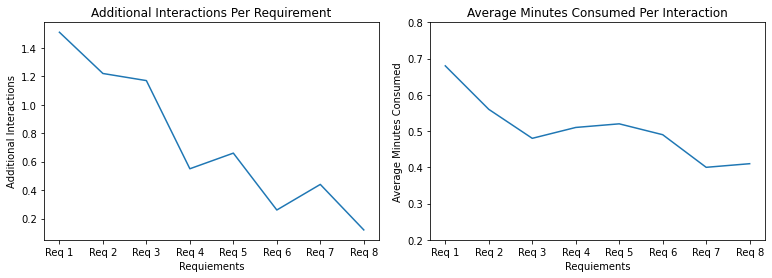}
    \caption{CitySpec in smart city  (case study)}
    \label{fig:smart_city_study}
    \vspace{-0.5cm}
\end{figure}

\newcontent{As Figure \ref{fig:smart_city_study} shows, there is an overall decrease in the additional interactions per requirement. This decrease demonstrates that our participants get almost as familiar as us with CitySpec. The slight increases at Req3, Req5, and Req7 are caused by the differences in participants' subjective understanding regarding the definition of key elements (e.g., $\mathsf{entity}$ or $\mathsf{description}$). Figure \ref{fig:smart_city_study} also shows the time consumed per interaction is decreasing as our participants use CitySpec. This decreasing trend indicates the participants are getting more familiar with those key elements with the help of CitySpec.}

\subsection{CitySpec in Unseen Domains}

\begin{figure}[t]
    \centering
    \includegraphics[width=0.98\textwidth]{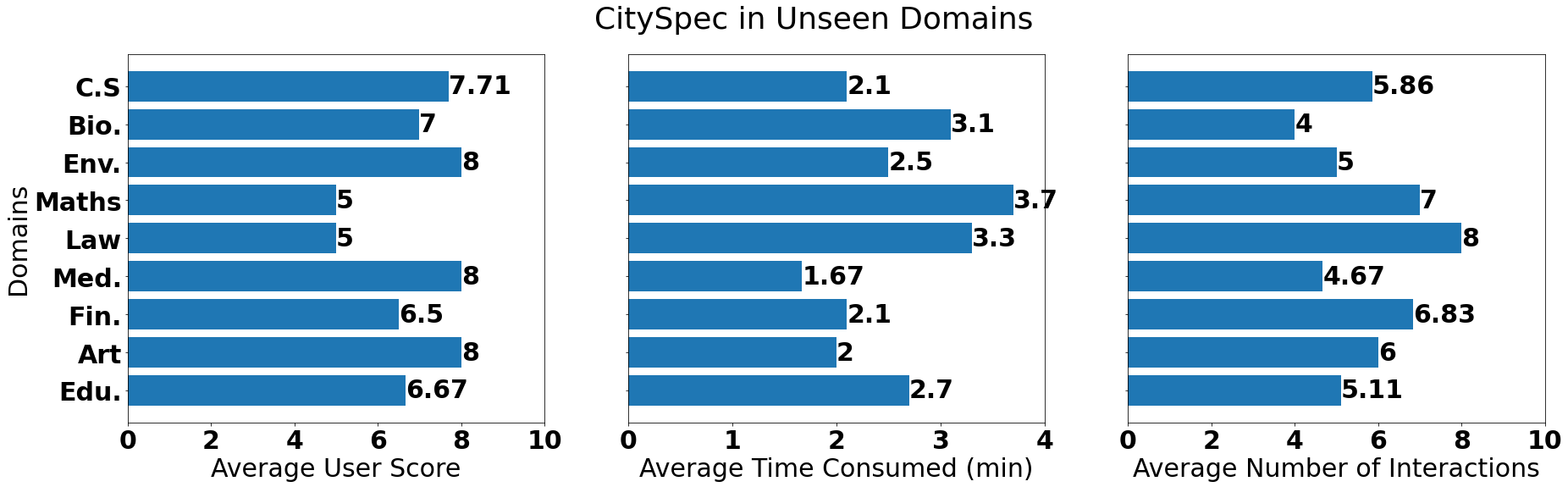}
    \caption{CitySpec in unseen domains (case study)}
    \label{fig:unseen_domain_study}
    \vspace{-0.5cm}
\end{figure}

\newcontent{To test the usability and adaptability of CitySpec in unseen domains, next, we ask the participants to compose three requirements regarding their own major and occupation. We record the average number of total interactions per requirement, the average user score, and the average time consumed per requirement. The results are shown in Figure \ref{fig:unseen_domain_study}. Among those provided requirements, CitySpec is found to work effectively in domains that are related to city regulation topics. For example, in the requirements from domains like Environmental Engineering, phrases like ``PM 2.5'' and ``iron concentration'' are passed to CitySpec. Although they are from a completely different area compared with city requirements, CitySpec still manages to correctly tell them from the requirements due to the partial similarity between city and environmental engineering.}

\newcontent{Although areas like medical, arts, and engineering are unseen during the training of CitySpec, CitySpec still offers correct results most times. For example, without learning ``acoustic reflexes'' as an entity during training, CitySpec presents the correct classification only based on the limited syntactic information given in the requirement. CitySpec's online learning feature also helps the specification. One participant whose major is Electronic Engineering composes requirements regarding system-on-chip/SoC and the circuit. In the beginning, CitySpec fails to tell terms like ``SoC'' and ``Circuit'' due to its unfamiliarity with them, however, after only a few turns of validated user clarification, CitySpec memorizes these unseen knowledge entries and gives correct classification in similar future cases. However, in areas like Law, the participant emphasizes the internal relationships between components and the `absolute' correct classification, e.g., including the determinator in the final prediction. Thus, the participant needs to continuously interact with CitySpec to generate the correct specification, which results in a relatively lower user score.}

\subsection{CitySpec's Defense Against Malicious Input}

\newcontent{To test the effectiveness of the newly introduced shield function, we ask the participants to provide both malicious and non-malicious inputs during CitySpec CitySpec's online learning in smart city. We collect 178 unique user clarification entries in total. Among those entries, 147 are considered malicious and 25 are considered non-malicious by the participants. With the help of our newly deployed shield function, CitySpec rejects 116 of those malicious inputs and yields a 0\% false-positive rate when dealing with non-malicious ones. Among those intentionally designed malicious inputs, two genre draws our attention: phonetic attacks and quantitative attacks. Here we define phonetic attacks as replacing words in given legit entries with their homophones but keeping the semantics shifted within a trustful range. Among all 11 phonetic attacks, 9 of them succeed. For example, the shield function fails to abort user clarification when ``fifteen'' is changed to ``fifty teen'', or ``plane ticket'' is changed to ``plain ticket''. Another genre is quantitative attacks, we define it as granting an unpractical quantity to components based on common senses, for example, assigning ``7k dB'' to ``the noise level at any personal property''.  Among 3 quantitative attacks, 3 of them succeed. Besides these two genres, CitySpec with the shield function reports 87.23\% as the defense success rate against all those malicious inputs.}

\newcontent{\textit{In summary, according to our real user case study, CitySpec works effectively and efficiently not only in smart city but also in most unseen areas like Environmental Engineering, Medical, and Arts, which also proves the potential future application of CitySpec in other areas. The decrease in both additional interactions per requirement and time consumed per requirement indicates the strong usability of CitySpec to users. The newly deployed shield function helps CitySpec be more robust to most malicious inputs.}}

\section{Related Work}
\label{sec:related}

\textbf{Translation Models.} Researchers have developed models to translate the natural language to machine languages in various applications, such as Bash commands~\cite{fu2021transformer}, Seq2SQL~\cite{zhong2017seq2sql}, and Python codes~\cite{chen2021evaluating}. These translation models benefit from enormous datasets. The codex was trained on a 159 GB dataset that contains over 100 billion tokens. WikiSQL, which Seq2SQL was trained on, consists of 80,654 pairs of English-SQL conversions. NL2Bash \cite{fu2021transformer} was trained on approximately 10,000 pairs of natural language tasks and their corresponding bash commands. 
As an under-exploited area, there is a very limited number of well-defined requirements. Therefore, existing translation models do not apply to our task. This paper develops a data synthesis-based approach to build the translation model. 


\textbf{Data Synthesis.} Data synthesis exploits the patterns in study findings and synthesizes variations based on those patterns. Data augmentation is a simple application of data synthesis. Previous augmentation approaches wield tricks like synonym substitution~\cite{kobayashi2018contextual, zhang2015character} and blended approaches~\cite{wei2019eda}. In the smart city scenario, we need new data samples which fit in the smart city context. Therefore, we extract extra knowledge from smart cities and fully exploit semantic and syntactic patterns instead of applying straightforward tricks like chopping, rotating, or zooming. This paper is the first work synthesizing smart-city-specific requirements to the best of our knowledge. 

\textbf{Online Learning.} Online machine learning
mainly deals with the situation when data comes available to the machine learning model sequentially after being deployed. Similar to continuous learning, online learning aims to give model accumulated knowledge and improve model performance continuously given incoming learning samples~\cite{parisi2019continual, chen2018lifelong}. Some of the existing papers focus on developing sophisticated optimization algorithms~\cite{hazan2016introduction} or exploiting the differences between new and old samples~\cite{sutton2018reinforcement}. However, these papers do not have a mechanism to detect or prevent adversarial samples online. This paper develops a two-stage online learning process with online validation against potential malicious inputs.  

\newcontent{\textbf{Protection against Adversarial Attacks.} Works have been done to enhance the safety of NLP models based on textual adversarial attacks. However, some of them make observations that pre-trained language models like BERT or GPT are robust to adversarial perturbations. For example, Bert-Defense \cite{keller-etal-2021-bert} uses Bert and GPT to lower the confusion and increase the fluency in generated samples; \cite{hendrycks-etal-2020-pretrained} uses pre-trained transformers to distinguish out-of-distribution samples. However, these pre-trained language models are trained on large corpora. Without a specific focus, those pre-trained models only offer limited prior knowledge in detecting malicious attacks in smart city. Other tricks like resampling are also widely applied and reported to succeed in ensuring system security \cite{rusert-srinivasan-2022-dont}. These methods still feed textual inputs to the backend model. Textual inputs give the attacker more room to tune with the input in our interactive and always-learning system. This paper develops a novel shield model that not only takes the prior knowledge from pre-trained models as references but also enriches the prior knowledge by introducing existing city requirements. No textual input is directly input to the shield model, instead, it encrypts the inputs so that it is considered immune to direct adversarial textual generation.}

\section{Summary}
\label{sec:summary}

This paper builds an intelligent assistant system, CitySpec, for requirement specification in smart cities. CitySpec bridges the gaps between city policy makers and the monitoring systems. It incorporates city knowledge into the requirement translation model and adapts to new cities and application domains through online validation and learning. The evaluation results on real-world city requirement datasets show that CitySpec is able to support policy makers accurately writing and refining their requirements and outperforms the baseline approaches. In future work, we plan to have CitySpec used by real city policy makers, but this is outside the scope of this paper.




\section*{Acknowledgment}
This work was funded, in part, by NSF CNS-1952096. 




 \bibliographystyle{elsarticle-num} 
 \bibliography{cas-refs}





\end{document}